\documentclass[lettersize,journal]{IEEEtran}
\usepackage{amsmath,amsfonts}
\usepackage{algorithmic}
\usepackage{algorithm}
\usepackage{array}
\usepackage[caption=false,font=normalsize,labelfont=sf,textfont=sf]{subfig}
\usepackage{textcomp}
\usepackage{stfloats}
\usepackage{url}
\usepackage{verbatim}
\usepackage{graphicx}
\usepackage{cite}
\usepackage{multirow}
\usepackage{booktabs}
\usepackage{color, xcolor}
\usepackage{enumitem}

\usepackage[
pdfauthor={derajan},
pdftitle={How to do this},
pdfstartview=XYZ,
bookmarks=true,
colorlinks=true,
linkcolor=blue,
urlcolor=blue,
citecolor=blue,
pdftex,
bookmarks=true,
linktocpage=true,   
hyperindex=true
]{hyperref}
\definecolor{yellow}{HTML}{E3A907}
\definecolor{green}{HTML}{4BA42C}
\definecolor{orange}{HTML}{C04F15}

\newcommand{\ar}[1]{{\color{black}#1}}

\begin{document}

\title{Explainable Action Form Assessment by Exploiting Multimodal Chain-of-Thoughts Reasoning}

\author{Mengshi Qi,~\IEEEmembership{Member,~IEEE,}
        Yeteng Wu,
        Wulian Yun,
        Xianlin Zhang,
        Huadong Ma,~\IEEEmembership{Fellow,~IEEE} 
\thanks{This work is partly supported by the Funds for the NSFC Project Grant 62572072, Beijing Natural Science Foundation (L243027). (\emph{Corresponding author: Corresponding author: Mengshi Qi and Xianlin Zhang~(email:~qms@bupt.edu.cn)})}
\thanks{M. Qi, Y. Wu, W. Yun, X. Zhang and H. Ma are with State Key Laboratory of Networking and Switching Technology, Beijing University of Posts and Telecommunications, China.}
}


\maketitle

\begin{abstract}

Evaluating whether human action is standard or not and providing reasonable feedback to improve action standardization is very crucial but challenging in real-world scenarios. However, current video understanding methods are mainly concerned with what and where the action is, which is unable to meet the requirements. Meanwhile, most of the existing datasets lack the labels indicating the degree of action standardization, and the action quality assessment datasets lack explainability and detailed feedback. Therefore, we define a new Human Action Form Assessment~(AFA) task, and introduce a new diverse dataset~\emph{CoT-AFA}, which contains a large scale of fitness and martial arts videos with multi-level annotations for comprehensive video analysis. We enrich the CoT-AFA dataset with a novel Chain-of-Thought explanation paradigm. Instead of offering isolated feedback, our explanations provide a complete reasoning process—from identifying an action step to analyzing its outcome and proposing a concrete solution. Furthermore, we propose a framework named Explainable Fitness Assessor, which can not only judge an action but also explain why and provide a solution. This framework employs two parallel processing streams and a dynamic gating mechanism to fuse visual and semantic information, thereby boosting its analytical capabilities.
The experimental results demonstrate that our method has achieved improvements in explanation generation (\emph{e.g.}, +16.0\% in CIDEr) ,action classification (+2.7\% in accuracy) and quality assessment (+2.1\% in accuracy), revealing great potential of CoT-AFA for future studies. Our dataset and source code are available at https://github.com/MICLAB-BUPT/EFA. 
\end{abstract}


\section{Introduction}
 \label{sec1}
\begin{figure}[t]
\begin{center}
 \includegraphics[width=\linewidth]{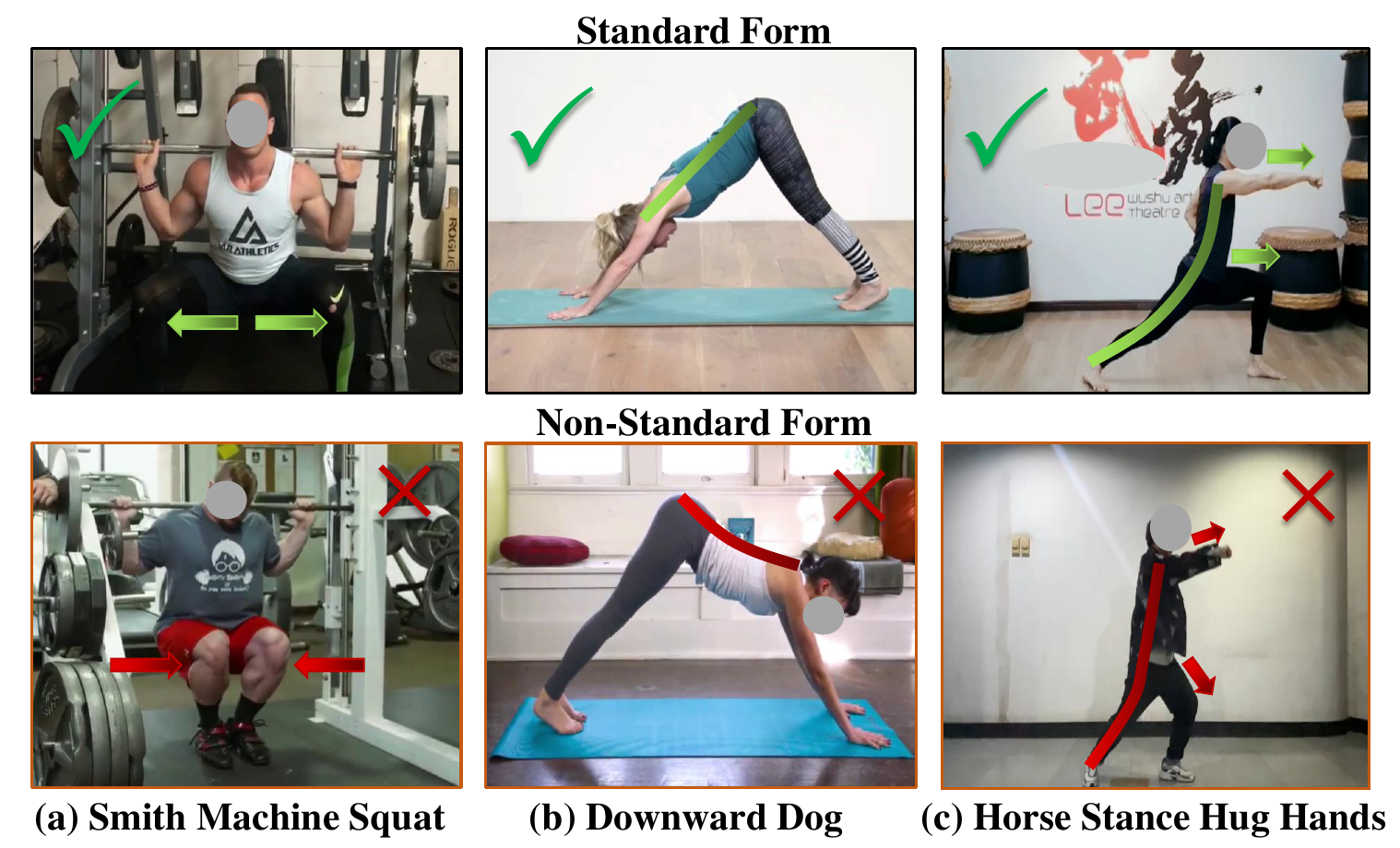}
\end{center}
\vspace{-4.5mm}
    \caption{Illustrations of standard and non-standard action examples in our proposed CoT-AFA dataset. The green and red lines indicate the correct and wrong actions, {respectively}. }
\label{fig:MATA}
\vspace{-4.5mm}
\end{figure}

\IEEEPARstart{I}{n} real-world scenarios, such as industrial production, healthy entertainment, and martial arts, it is crucial to \ar{assess whether the action is performed in a standard form}. Therefore, we introduce a new task, \emph{i.e.}, \emph{Human Action Form Assessment~(AFA)}. Different from existing video understanding tasks, such as action recognition ~\cite{10.5555/2968826.2968890,8454294,7558228,9008827,Carreira_2017_CVPR,9008780}
and localization~\cite{8237579,Huang_2022_CVPR,he2022asm}, AFA moves beyond merely identifying what or where an action is performed. Instead, it focuses critically on \emph{how well} the action conforms to pre-defined objective standards, as illustrated in Figure \ref{fig:MATA}. This assessment is vital for ensuring effectiveness, safety, and performance optimization in various domains.

In recent years, there have been significant advances in video datasets for understanding human action, such as Kinetics~\cite{articleKinetics} and UCF101~\cite{Soomro2012UCF101AD} for video classification, ActivityNet~\cite{7298698} and THUMOS~\cite{THU} for temporal action localization. However, they mainly provide annotations for the action categories or timestamps observed in the videos, without annotations about action quality.
Although several emerging action quality assessment~(AQA) datasets~\cite{10.1007/978-3-319-10599-4_36,Parmar2018ActionQA} record the action scores to reflect the quality of the actions, they are mainly derived from subjective evaluations by individuals and only applied to sports scenarios, which entail certain limitations. In real-life situations, objective standards should be used to assess whether the action is standard form or not. Therefore, these existing datasets are not applicable to AFA. 

To bridge this gap and establish a foundation for the AFA task, we introduce a new diverse video dataset named \emph{CoT-AFA}, focusing on a wide range of workout actions, including fitness and martial arts. We chose the workout domain for several strategic reasons: workouts encompass a diverse range of movements targeting different muscle groups; they possess clear, objective, and quantifiable standards, facilitating unambiguous annotation of standard and non-standard forms; and the applications are broad, spanning from personalized coaching and rehabilitation to AR/VR training simulations.
Finally, the workout dataset has a wide range of application scenarios, such as fitness teaching, martial arts teaching, action counting and AR/VR action generation, and can also be extended to other fields, \emph{e.g.}, manufacturing instruction in industry.

However, a simple binary (standard/non-standard) judgment is often insufficient for real-world utility. For effective personalized coaching or skill improvement, users require not only a verdict but also actionable, explanatory feedback. Traditional instructional systems often provide isolated, generalized tips \cite{panchal2024say}  (\emph{e.g.}, ``keep your back straight'') that fail to address the complex causal link between a movement error and its consequence.
Inspired by the effectiveness of Chain-of-Thought~(COT)  reasoning in large language models (LLMs) \cite{wei2022chain}, we enrich the CoT-AFA dataset with a novel CoT explanation paradigm. Our annotations transition from isolated feedback to a structured reasoning process. This process begins with identifying the specific non-standard action step, and then analyzes the observable outcome (\emph{e.g.}, ``This poor form will cause excessive strain on the shoulders''), and concludes by proposing a concrete solution. This multi-stage, causal explanation mirrors the diagnostic process of a human expert, significantly enhancing the interpretability and utility of the assessment.


To effectively utilize the structured information within CoT-AFA, we propose a novel Explainable Fitness Assessor (EFA) framework. EFA is designed to perform comprehensive action evaluation by leveraging predefined standard technical steps as the explicit structured guidance. The core idea is to move beyond passive feature extraction by actively aligning visual evidence with textual instructions. The framework employs a hierarchical multi-modal architecture: 1) Dual Parallel Processing: EFA utilizes two parallel processing streams, including a Step-Aware Fusion Module, which grounds fine-grained visual details within the context of specific procedural steps, and a Global-Aware Fusion Module, which aligns video features with the overall action goal. 2) Bidirectional Attention: Leveraging a bidirectional attention mechanism, ensuring that visual features are contextually informed by the text, and simultaneously, the text is temporally grounded in the video's most relevant segments. 3) \ar{Dynamic Gating Mechanism: Following parallel processing, a dynamic gating fusion layer adaptively integrates the outputs from both step-aware and global-aware streams. Unlike traditional CLIP-based paradigms or vanilla fusion mechanisms that rely on flat, sequence-level feature aggregation for coarse semantic matching, our proposed dynamic gating explicitly addresses the strict causal logic required for fine-grained action assessment. By explicitly calculating a frame-level gating weight through a dedicated linear mapping and Sigmoid activation on the concatenated dual-branch features, this mechanism allows the model to dynamically prioritize the most influential level of detail for the final judgment.} Finally, the fused features are used to simultaneously perform action classification, action quality assessment, and generate the detailed Chain-of-Thought textual explanation.

Our contributions in this paper are highlighted as follows: 
\begin{itemize}
\item We construct a new video dataset CoT-AFA to fit into the new human action form assessment task, which contains more than 364k frames and 3k rich multi-element annotations to enhance the interpretability of action assessment. 

\item We propose a novel Explainable Fitness Assessor framework that utilizes predefined standard technical steps to learn quality features and perform action assessment.

\item
Extensive experiments illustrate that our proposed method obtains substantial improvements and conducted comprehensive studies indicate that the challenges of CoT-AFA and great potential of the AFA task.
\end{itemize}

\begin{figure*}[t]
\begin{center}
\setlength{\fboxrule}{0pt}
\hspace{0mm}
\includegraphics[width=0.9\linewidth]{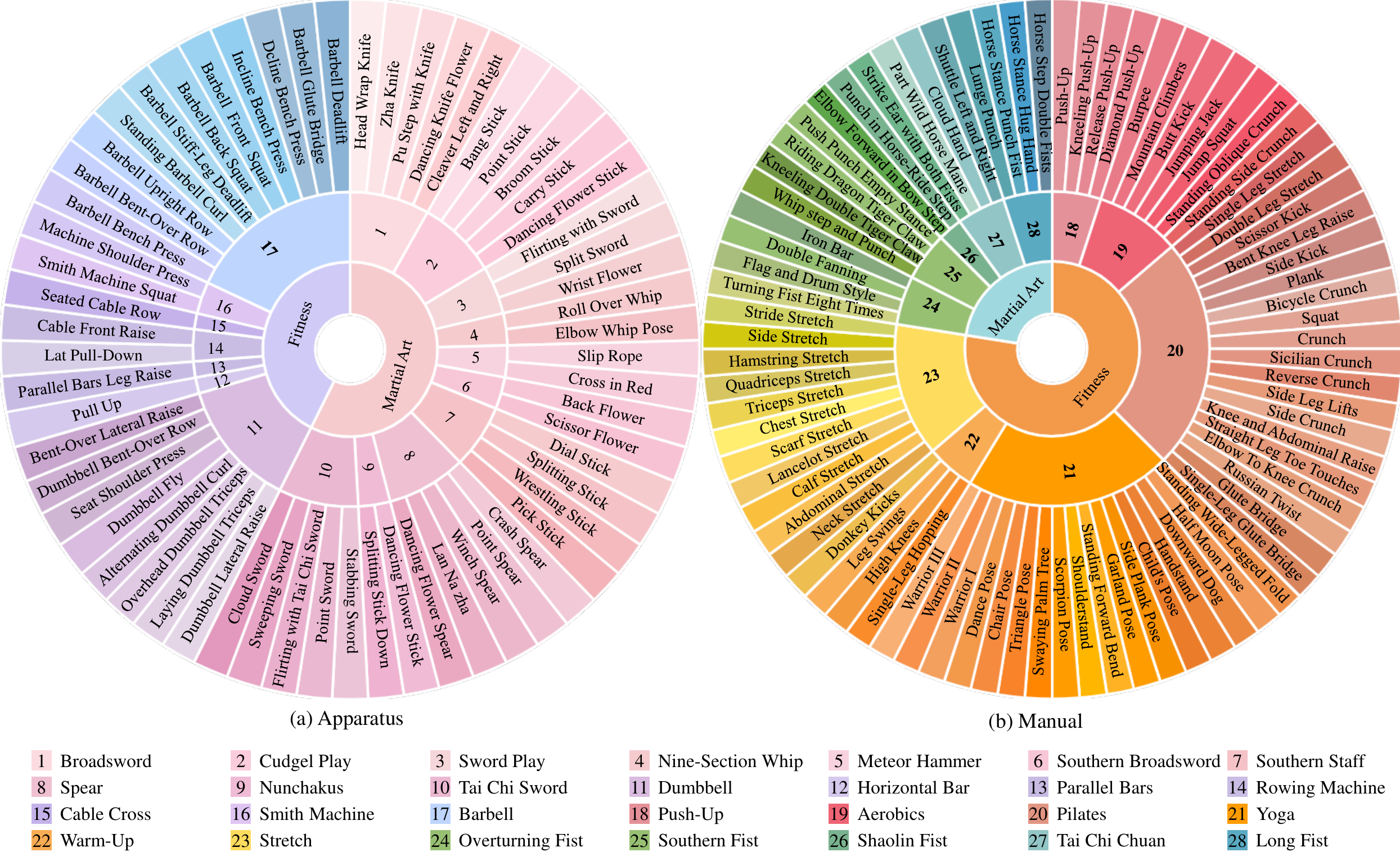}
\end{center}
\vspace{-4.5mm}
    \caption{Three-level lexicon annotation structure of apparatus~(left) and manual~(right). The first colored layer outside the center of the circle represents martial arts and fitness. The next layer represents the workout type. The outermost layer refers to the action category. }
\vspace{-3mm}
\label{fig:2}
\end{figure*}

\begin{figure*}[t]
\begin{center}
     \includegraphics[width=0.95\linewidth]{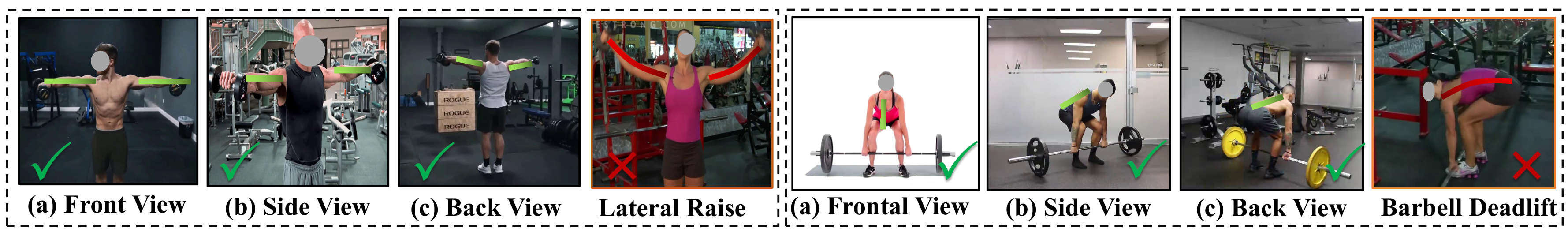}
\end{center}
\vspace{-5mm}
    \caption{Illustration of multi-element annotations of  CoT-AFA dataset, which shows two categories of actions from different views, including front view, side view, and back view, with both standard and non-standard samples for each action. }
\vspace{-3mm}
\label{fig:3}
\end{figure*}

\section{RELATED WORK}
 \label{sec2}
\noindent

\noindent\textbf{Video Understanding.}
Video understanding is always a challenging problem in the field of computer vision and multimedia, which aims to enable computers to interpret and analyze video content. It contains many subtasks including action 
recognition~\cite{10.5555/2968826.2968890,8454294,7558228,9008827,zhou2017temporalrelation,Carreira_2017_CVPR,9008780,10.5555/3504035.3504947,cheronICCV15,
qian2025beyond,zhu2025semantic,qi2019attentive,qi2020stc,lv2025t2sg,zhu2023unsupervised}, action localization~\cite{8237579,Huang_2022_CVPR,he2022asm,Lin_2018_ECCV,zhang2025weakly,lv2023disentangled,qi2018stagnet,qi2020imitative,qi2019ke,qi2025robust,11223230}, etc. 
Most widely-used action recognition methods adopt the two-stream structure~\cite{10.5555/2968826.2968890,8454294,7558228,9008827,zhou2017temporalrelation} to model both spatial and temporal information. Meanwhile, the transformer-based methods can be treated as the third category, such as Video Swin~\cite{Liu_2022_CVPR}, Timesformer~\cite{gberta_2021_ICML}, MViT v2~\cite{li2021improved}, etc. Another key research avenue involves skeleton-based approaches, which are also proposed to address the issue~\cite{10.5555/3504035.3504947,cheronICCV15}.
More recently, the field has shown a growing trend towards incorporating additional modalities, with text being a prominent example. Illustrating this, Qian~\emph{et al.}~\cite{qian2025beyond} proposed a language-guided approach to action decomposition that resolves an action label into an ordered sequence of descriptions. In a similar vein, Zhu~\emph{et al.}~\cite{zhu2025semantic} leveraged bimodal data from text and skeletons to learn prompts. Furthermore, Zhang~\emph{et al.}~\cite{zhang2025weakly} harnessed the semantic priors from Multimodal Large Language Models  to provide guidance for temporal action localization models. \ar{More broadly, while MLLMs have remarkably advanced video understanding \cite{tang2025video,zhang2024mm,kuang2025natural,song2025bridge,qin2025synergy,wu2025survey}, they primarily excel at action classification and localization. Their inability to evaluate whether a movement is executed correctly based on objective criteria is exactly why we introduce our proposed AFA task.}

\noindent
\textbf{Action Quality Assessment.}
The main purpose of AQA is to determine \emph{how well} the action is performed. Current existing methods~\cite{10.1007/978-3-319-10599-4_36,Gordon1997AutomatedVA,8014750,Parmar_2019_CVPR,8578732,10.1007/978-3-031-19772-7_27,11185304,yun2024semi,
wang2025learning,an2024multi,ye2025safedriverag}
formulate the AQA as a regression task, by employing scores as supervised signals. For example, Pirisiavash~\emph{et al.}~\cite{10.1007/978-3-319-10599-4_36} utilized the Support Vector Machines model~\cite{10.5555/2998981.2999003} to map pose features to video scores. Gordon~\emph{et al.}~\cite{Gordon1997AutomatedVA} assessed the quality of gymnastic vaulting movements using skeleton trajectories. 
Parmar~\emph{et al.}~\cite{8014750} employed 3D convolutional neural networks (CNNs) to extract spatio-temporal features and enhance the performance of AQA. 
Li~\emph{et al.}~\cite{10.1007/978-3-031-19772-7_27} used pairwise comparison learning methods to capture differences between videos while simultaneously constructing separate branches to learn relative scores. Wang~\emph{et al.}~\cite{wang2025learning} proposed a dual-stream Mamba pyramid network for video action assessment. Several recent studies have started focusing on generating evaluative texts. Such methods usually adopted an encoder-decoder architecture, mapping video features to corrective or evaluative descriptions. Zhang~\emph{et al.}~\cite{zhang2024narrative,panchal2024say} introduced a prompt-guided multimodal framework that transforms score regression into a video-text matching problem, enabling the generation of detailed narrative evaluations of actions. The difference between AQA and AFA is that AFA needs to \ar{justify \emph{whether human actions are standard}}, while AQA is to estimate \emph{how well} the actions are performed. Moreover, scores used in AQA are subjective, while AFA requires an absolutely objective standard. Therefore, existing AQA methods and datasets are not suitable for the AFA task.


\noindent
\textbf{Video Captioning.}
This is a fundamental task with wide-ranging applications in computer vision. Some existing works~\cite{yu2018fine,qi2019sports,kaur2021automatic,qi2021semantics,lv2024sgformer} applied video captioning to the sports domain. Earlier approaches 
\cite{chen2019deep,lin2022swinbert,wang2021end,yang2023vid2seq,aafaq2019spatio} commonly focused on generating concise textual descriptions for input video sequences. For example, Lin~\emph{et al.} \cite{lin2022swinbert} proposed an end-to-end solution to perform visual feature extraction via Video Swin Transformer \cite{Liu_2022_CVPR}, and utilized a Transformer-based module \cite{vaswani2017attention} to generate the corresponding text sequences. Wu~\emph{et al.} \cite{wu2025event} proposed an event-equalized framework that leverages clustering on visual features to unbiasedly identify potential pseudo-events, thereby guiding the model to pay equal attention to events of all durations. Xi~\emph{et al.} \cite{xi2025player} integrated player identity features and names extracted directly from visual content with contextual video information to guide the LLM in producing player-centric captions. However, this type of caption generation falls short for fine-grained tasks such as action guidance. In contrast, our proposed Explainable Fitness Assessor method generates detailed, instructive explanations by combining standard action steps with quality judgment criteria, simulating expert reasoning.

\section{The CoT-AFA Dataset}
    \label{sec3}

To effectively address the novel task of AFA, we meticulously construct a new dataset called the CoT-AFA Dataset. Existing datasets for AQA primarily focus on providing numerical scores or coarse-grained labels for action proficiency \cite{Parmar_2019_CVPR,shao2020finegym}. However, they lack the detailed, step-by-step texts and causal reasoning necessary for providing actionable Chain-of-Thought explanation, which is the core requirement of our proposed task. The CoT-AFA dataset is specifically designed to bridge this gap by incorporating rich, structured explanations that mimic human expert evaluation processes.
In this section, we will introduce the CoT-AFA dataset from data preparation, data annotation, dataset statistics, and Chain-of-Thought Text Explanation.

\begin{table*}[h]
\centering
\caption{Statistics of our proposed CoT-AFA dataset. }
\renewcommand{\arraystretch}{1.1}
\vspace{-2mm}
\setlength{\tabcolsep}{1.0mm}{ 
\begin{tabular}{lcccccccc}
\specialrule{0.1em}{1pt}{1pt}
\hline
Dataset & Workout modes & Workout types & Action categories & Standard Videos & Non-standard Videos & Total Videos & Total Frames & CoT Text Explanations \\ \hline
CoT-AFA & 2 & 28 & 141 & 2,242 & 1,150 & 3,392 & 364,812 & 3,392 \\ \hline
\specialrule{0.1em}{1pt}{1pt}
\end{tabular}}
\vspace{-1mm}
\label{tab:1}
\end{table*}

\begin{table}[h]
\centering
\caption{Quantitative Statistics of Chain-of-Thought Text Explanation in CoT-AFA.}
\renewcommand{\arraystretch}{1.2}
\vspace{-2mm}
\setlength{\tabcolsep}{1.2mm}{
\begin{tabular}{lcc}
\specialrule{0.1em}{1pt}{1pt}
\hline
\textbf{Metric Category} & \textbf{Statistic} & \textbf{Value} \\ 
\hline
\multicolumn{3}{l}{\textbf{Text Scale and Volume}} \\

Average Words per Sample & Word Count & 102.19 \\ 
Average Sentences per Sample & Sentence Count & 5.25 \\
\hline
\multicolumn{3}{l}{\textbf{Linguistic Complexity}} \\

Total Vocabulary Size & Token Count & 3,143 \\
\hline
\multicolumn{3}{l}{\textbf{Chain-of-Thought Characteristics}} \\

Average Reasoning Steps (Logic) & Per Sample & 0.91 \\
Average Actionable Suggestions (Utility) & Per Sample & 0.75 \\
\hline
\specialrule{0.1em}{1pt}{1pt}
\end{tabular}}
\label{tab:text_stats}
\vspace{-2mm}

\end{table}

\begin{table}[h]
\centering
\caption{Comparison with existing datasets and CoT-AFA. Where `Cla' is class, `SC' denotes standardization category, and `M' represents multiple viewpoints. $*$ denotes action recognition dataset, and others represent the AQA dataset.}
\vspace{-2mm}
\renewcommand{\arraystretch}{1.2}
\setlength{\tabcolsep}{2.5mm}
\begin{tabular}{lccccc}
\specialrule{0.1em}{1pt}{1pt}
\hline
Dataset          & Category  & Time  & Samples & Labels \\ \hline
Finegym*~\cite{shao2020finegym}                    & 10             & 1.7 s       &  32,697    &   Cla     \\
HMDB51*~\cite{6126543}                     & 51             & 5.0 s       &  6,849    &   Cla    \\
MIT Dive~\cite{10.1007/978-3-319-10599-4_36}       &  1             & 6.0 s      &  159      &   Score     \\
UNLV-Dive~\cite{8014750}                          &  1             & 3.8  s    &  370      &   Score     \\
AQA-7~\cite{Parmar2018ActionQA}                    &  6             & 4.1  s    &  549      &   Score     \\
MTL-AQA~\cite{Parmar_2019_CVPR}                   & 16              & 4.1  s    & 1,412      &  Cla, Score    \\ 
FineDiving~\cite{Xu_2022_CVPR}                    & 30              &  4.2 s    & 3,000      &  Cla, Score    \\ 
WF~\cite{parmar2022win}                     &  4             & 3.3 s     &  1,643     &   SC   \\\hline
CoT-AFA                                          & 141              &  3.5 s     & 3,392     &  Cla, SC, M    \\ \hline
\specialrule{0.1em}{1pt}{1pt}
\vspace{-2mm}
\end{tabular}
\vspace{-5mm}
\label{tab:2}
\end{table}

\subsection{Data Preparation}
Our data collection procedure consisted of the following steps.
We started by identifying the categories of all workout actions, and we surveyed workout videos and actions with high popularity from various websites and fitness apps. Then, we select 141 types of workout actions to construct our dataset considering the popularity, usefulness (\emph{e.g.}, effectiveness in training the body's core muscles), diversity, and feasibility, thus ensuring that our dataset is a diverse and representative one. Afterwards, we collect relevant videos from \emph{YouTube} for each category with great completeness and high resolution (\emph{i.e.,} 720P and 1080P) during collection. Additionally, as our main task is to assess human action form, we collect data including both \emph{standard form} and \emph{non-standard form} action for each category. The definition of \emph{non-standard} is set by consulting fitness trainers as well as conducting extensive related research, involving the wrong movement of certain human body part. However, non-standard actions are still limited and too difficult to obtain through regular channels. Therefore, we enrich the dataset by shooting the bad form action video clips by own based on common errors during the execution of the workout. Finally, we manually edit the videos to ensure that only the parts related to the workout actions are preserved, which removes irrelevant information and protect personal privacy, leading to improve the data quality and facilitate the subsequent workout analysis.

\subsection{Data Annotation}
We perform rich multi-element annotations on the CoT-AFA dataset, including lexicon, duration and viewpoints. The details are as follows:

$\bullet$ \textbf{Lexicon.} 
1) We first define the lexicon of the CoT-AFA dataset in terms of three levels: \emph{workout mode}, \emph{workout type}, and \emph{action category}, detailed as shown in Figure~\ref{fig:2}. For the first level, we ask the annotator to classify and then annotate each workout into two modes,~\emph{i.e.}, manual and apparatus. Meanwhile, annotate whether the current action belongs to martial arts or fitness. The second level is to annotate the specific workout type based on different mode,~\emph{e.g.}, the apparatus contains the dumbbell or barbell while the manual includes aerobics or yoga. 
The third level is the specific action category corresponding to the second level, \emph{e.g.}, aerobics contains bobby jump, etc. The hierarchical lexicon and annotation make our dataset suitable for further action understanding.
2) 
Importantly, to achieve the action form assessment task, we request annotators to categorize each data as either \emph{standard} or \emph{non-standard} and provide a reason for each \emph{non-standard} action. As shown in Figure~\ref{fig:3}, the standard form of dumbbell lateral raise is to raise the dumbbells laterally to shoulder height with a slight bend in the arms. Raising the dumbbells too high or bending the arms to bring them level with the head are annotated as bad form, which can cause excessive strain on the shoulders and increase the risk of injury. Another example is the barbell deadlift, the standard form is to keep the back straight rather than bending over to lift. 

$\bullet$ \textbf{Duration and frame number.}
Due to variations in the video speed and length, we annotate each video with its duration and frame number, which facilitates statistical analysis of the dataset and is beneficial to subsequent pre-processing operations. In total, the dataset has 3,392 videos with 364,812 frames.

$\bullet$ \textbf{Multiple viewpoints.}
Since our video data is retrieved from Youtube that generally filmed by different individuals from multi-view cameras, we annotate the recording viewpoint of each video as \emph{front view}, \emph{side view}, and \emph{back view}, as shown in  Figure~\ref{fig:3}. Different views can offer more complementary information, which can enhance the accuracy and generalizability of subsequent recognition and assessment tasks, such as multi-view pose estimation and multi-view action recognition.

\begin{figure*}[t]
    \centering
         \includegraphics[width=0.8\linewidth]{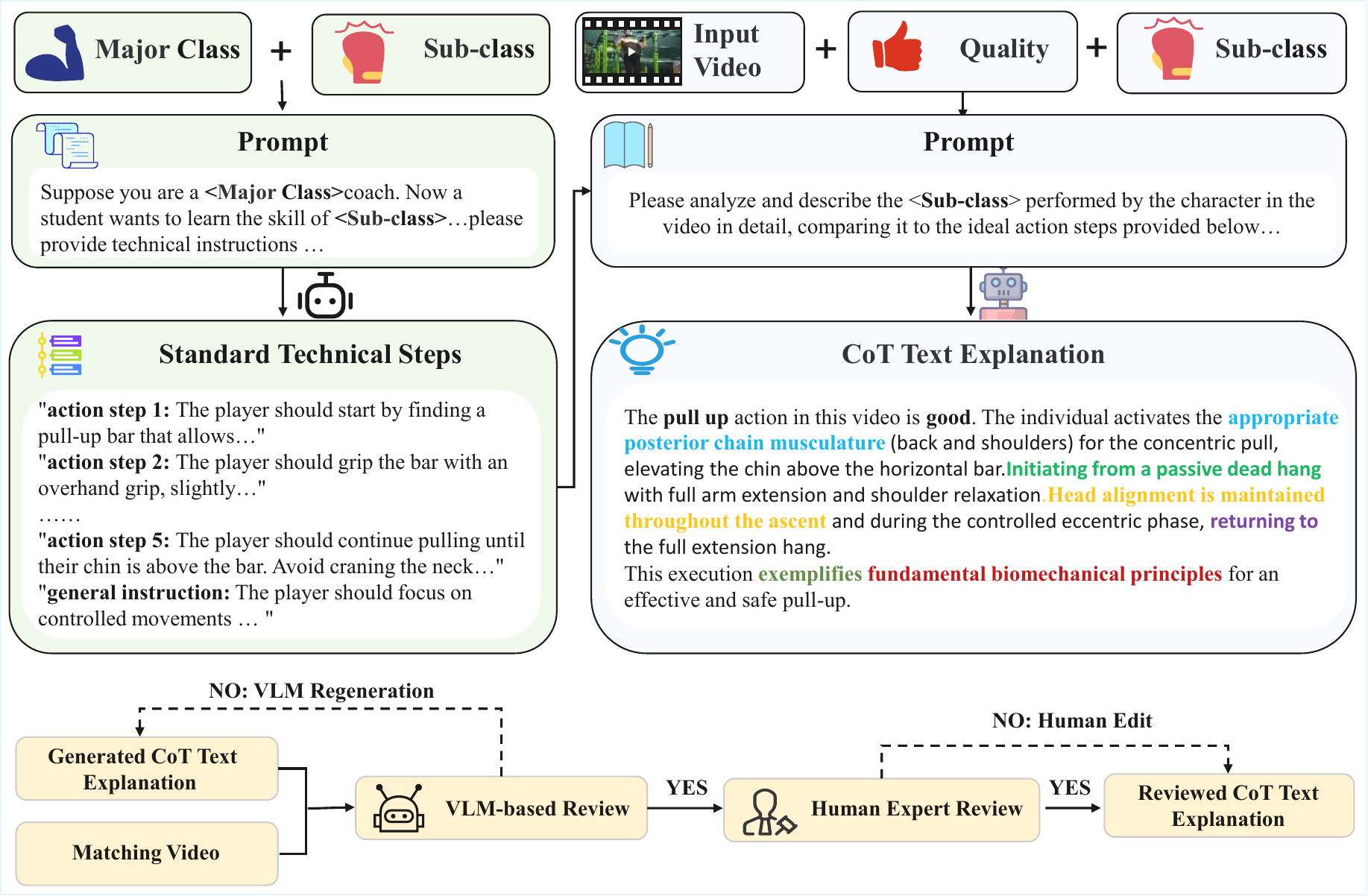}
         		\vspace{-1mm}
    \caption{The workflow for textual explanation generation process of CoT-AFA, showing the generation of Standard Technical Steps for each action using LLM and Chain-of-Thought Text Explanations by VLM, followed by a VLM-based and human expert review process to ensure quality.}
		\label{figure2}
		\vspace{-1mm}
	\end{figure*}

\subsection{Dataset Statistics}
Table~\ref{tab:1} shows the statistics of the CoT-AFA dataset. Specifically, our dataset contains two workout modes, 28 workout types and 141 action categories, consisting of 3392 video samples with 2242 standard and 1150 non-standard samples in total. This statistical analysis can help us better understand the characteristics of the dataset, thereby facilitating data preprocessing, feature extraction, and model training. Meanwhile, we also compare existing widely-used datasets with CoT-AFA in Table~\ref{tab:2}. We can find that current datasets only provide action categories or subjective score annotations, lacking standardization category annotations. Moreover, CoT-AFA dataset covers more action categories and rich annotations than others, making it best for the AFA task.

\subsection{Chain-of-Thought Text Explanation }
Our dataset introduces a novel explanation paradigm: Chain-of-Thought text explanations for human actions in videos. Unlike traditional methods that offer simplistic standard/non-standard verdicts or isolated behavioral suggestions, our approach provides a comprehensive analysis of action execution. It elucidates the ``why'' behind an action's assessment by uncovering complex causal relationships between steps, identifying underlying reasons for errors, and delivering actionable, corrective explanations.

As depicted in Figure \ref{figure2}, this sophisticated explanation is generated through a four-step pipeline:
First, we engineer a structured prompt template designed for broad applicability across our comprehensive set of action categories. Subsequently, this template is programmatically combined with a sub-class and fed into an LLM (\emph{i.e.,} Gemini 2.0). Following this, the LLM produces a set of Standard Technical Steps. Next, these generated steps are consolidated with the target action video and its pre-annotated quality label, which are then collectively input into a VLM~(\ar{\emph{i.e.,} VideoChat}). Ultimately, the VLM generates a detailed CoT text explanation.

To guarantee the quality and accuracy of these automatically generated explanations, the data then undergoes a rigorous two-stage review process, As depicted in Figure \ref{figure2}. First, we employ a powerful VLM (\emph{i.e.,} Qwen2.5-VL), to conduct an automated logical consistency check. This step serves to evaluate the logical consistency of the generated text and its correspondence with the video content.
Subsequently, each sample is passed to a team of eight human experts proficient in workout for a meticulous manual review. These experts perform a multi-faceted evaluation: they verify the correctness of the assigned action category and quality label, and critically, they ensure a high-fidelity correspondence between the textual explanation and the specific, fine-grained details of the action performed in the video. Any generated text that fails this thorough vetting process is flagged and manually edited by the experts to ensure the final dataset meets our stringent quality standards.

To deeply investigate the characteristics of the text explanations in our collected CoT-AFA dataset, we conducted a series of quantitative statistical analyses. These analyses cover the basic scale, linguistic complexity, and the core Chain-of-Thought structure of the text. The statistical results, as shown in Table \ref{tab:text_stats} comprehensively demonstrate the dataset's richness and high quality.
Regarding text scale, the average length of the explanation text per sample is approximately 102.19 words, consisting of an average of 5.25 sentences. This ensures that each explanation contains sufficient detail and context, avoiding oversimplified descriptions.
In terms of linguistic complexity, the entire dataset has a vocabulary size of 3,143, demonstrating good linguistic diversity and coverage of professional terminology.
Crucially, to quantify the text's CoT properties and practical utility, we assessed both the reasoning logic and the actionable suggestions:
1) Reasoning Logic: Each sample includes an average of 0.91 reasoning steps (identified through causal and transitional keywords such as ``because'', ``therefore'' and ``as a result''). This indicates that the explanations go beyond simple phenomenological descriptions, incorporating a causal analysis that links observed errors to their underlying causes.
2) Practical Utility: Each sample provides an average of 0.75 actionable suggestions (identified through corrective keywords and structures). This highlights the dataset's value in offering specific, practical, and actionable feedback that helps users understand how to correct their movements.
Collectively, these statistical features confirm the high quality of the CoT-AFA dataset in terms of content depth and structural logic, providing a solid foundation for future research in AFA.

\section{Methodology}
We first present the problem definition of action form assessment and then describe our proposed approach. 

\textbf{Problem Formulation.}
The action form assessment task aims to assess whether standard the action was performed and generate explanatory feedback. In our method, 
given an input video $ {V}$ and a set of predefined standard technical steps $ {T}$ for that action's class, our goal is to train an action form assessment model to predict a tuple~($ {y}_{{c}}$, $ {y}_{{q}}$, $ {t}_{{c}}$). Here, $ {y}_{{c}}$ is the predicted action class, $ {y}_{{q}}$ is the predicted quality label, and $ {t}_{{c}}$ is the generated Chain-of-Thought textual explanation.

\begin{figure*}[t]
    \centering
    \includegraphics[width=0.9\linewidth]{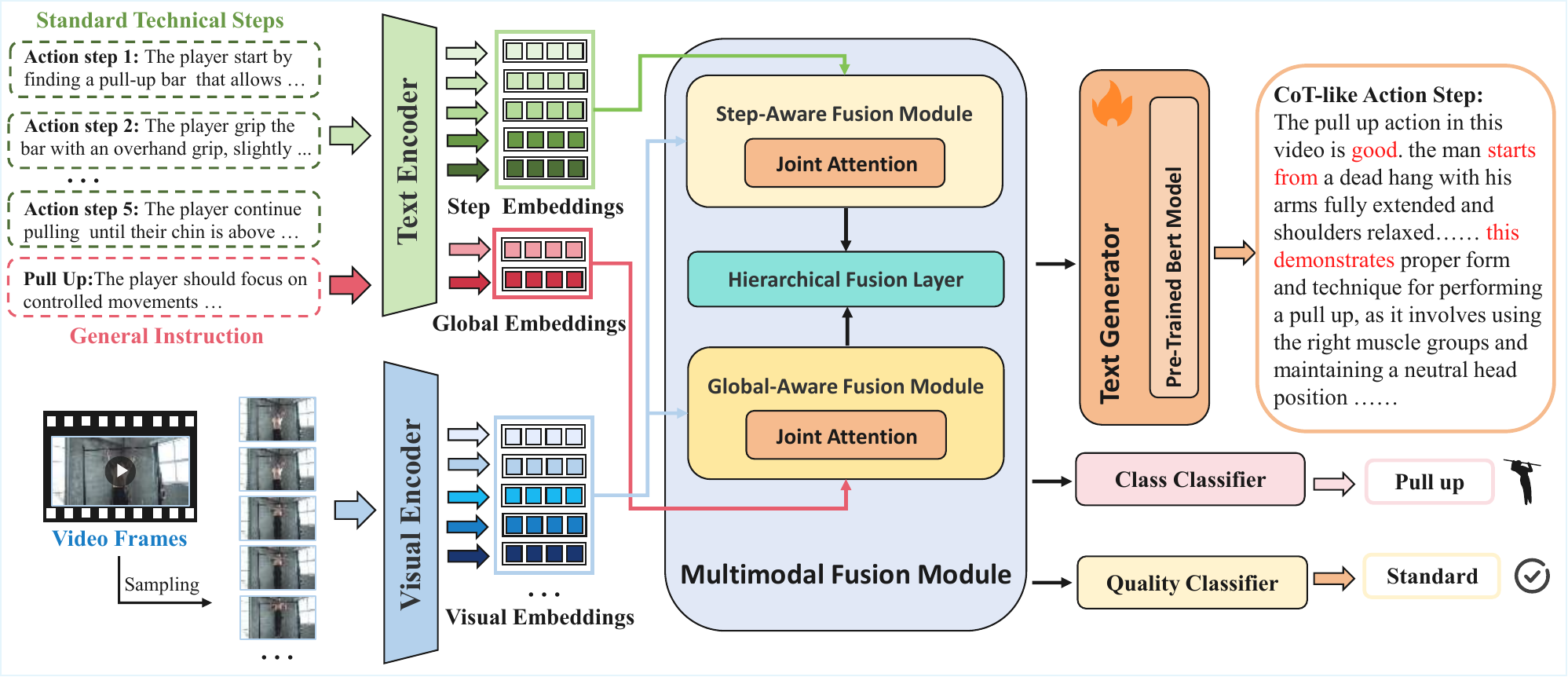}
    \vspace{-2mm}
    \caption{The architecture of our proposed Explainable Fitness Assessor~(EFA). EFA receives video frames and text as input. Visual and text features are extracted by the backbone. The Multimodal Fusion Module is used to fuse visual and text information. Finally, the framework outputs the action quality score, action class and Chain-of-Thought text explanations.}
		\label{figure1}
		\vspace{-2mm}
	\end{figure*}

\textbf{Overview.}
The overall architecture of our method is illustrated in Figure \ref{figure1}.
Our method evaluates action quality and generates instructional feedback through a multi-level framework. First, we input the video into the visual encoder to extract visual features. Then, a set of standard technical steps corresponding to the video action category is processed using a text encoder and converted into embedding vectors, which represent the semantic meaning of several action steps and a general instruction in the form of high-dimensional vectors. To fuse multimodal information at different levels and granularities, we design a new Multimodal Fusion Module with two parallel branches: the Global-Aware Fusion Module, which fuses visual embeddings with global embeddings, and the Step-Aware Fusion Module, which fuses visual embeddings with step embeddings. Then a Hierarchical Fusion Layer integrates the outputs from these two modules and creates a fused representation by aligning two modalities.  Finally, these fused features are passed to separate prediction heads to simultaneously predict the action class being performed in the video, assess the action quality and generate the corresponding Chain-of-Thought textual explanation.

\subsection{Visual and Text Encoder}    
Given an input video $ {V} $, it is first sampled into a sequence of raw frames
$ {f}_{{1}}$, $ {f}_{{2}}$, $ {f}_{{3}}$ ... $ {f}_{{T}}$, where each frame has dimensions $ {f} \in \mathbb{R}^{H \times W \times 3}$. 
This preprocessed video sequence is fed into a pre-trained Video Swin Transformer \cite{Liu_2022_CVPR}. 
The Video Swin Transformer processes this input tensor and outputs a sequence of compressed and semantically rich video feature tokens $ {\mathcal{F}_{{v}}} \in \mathbb{R}^{{N} \times {D}}$. Here, $ {N} $
 represents the total number of spatio-temporal tokens generated by the transformer, and 
$ {D} $ is the dimensionality of the feature vector for each token. 

Given the input set of standard technical steps ${T}$ that corresponds to the action class in the video, it is fed into a pre-trained text encoding model~(\emph{e.g.}, BERT-base-uncased \cite{devlin2019bert}) to process the input text. To describe the action from multiple levels of granularity,  $ {T} $   comprises two core components. The first component consists of the ideal  five action steps defined for each action category, representing a micro-level, fine-grained description of the execution process, while the second component is a general instruction, representing a global summary of the action. Both of these textual components are fed into the shared text encoder to generate a series of step embeddings, $ {\mathcal{F}}_{{t}}^{s} \in \mathbb{R}^{{M} \times {E}}$ , and global embeddings $ {\mathcal{F}}_{{t}}^{g} \in \mathbb{R}^{1 \times {E}}$, where $ {M} $ represents the number of text steps, and $ {E} $ is dimensionality of embedding vector for each step. \ar{Specifically, we directly utilize the raw outputs from the BERT encoder, without introducing any additional linear projection heads. This streamlined design avoids redundant parameters, as the subsequent cross-attention modules inherently perform spatial-semantic projections ($W_Q$, $W_K$, $W_V$), while maximally preserving the pre-trained semantic priors.}


\subsection{Multimodal Fusion Module}
To effectively fuse information from different levels of abstraction and granularity, our Multimodal Fusion Module is designed with a hierarchical architecture. It processes features in two parallel, specialized branches before integrating their outputs. 

\subsubsection{Global-Aware Fusion Module} 
As shown in Figure \ref{figure_1}, this module is responsible for enriching the visual features with global embeddings $ {\mathcal{F}}_{{t}}^{g}$. Inspired by previous work \cite{zhang2024narrative}, to achieve bidirectional cross-modal information interaction, we partition this module into two parts for learning.

Initially, to enable the text features to focus on the key video parts that match specific action instructions, we employ the global embeddings $ {\mathcal{F}}_{{t}}^{g}$  as queries $ {Q}_{{t}}^{{g}}$ and the video features $ \mathcal{F}_{{v}}$ as keys $ {K}_{{v}}$ and values $ {V}_{{v}}$. This process can be represented as:
\begin{equation}  
         {\mathcal{F}}_{{t}}^{{g*}}= Attn_1 (\mathcal{F}_{t}^{g}, \mathcal{F}_{v})+\sigma _{{1}}{\mathcal{F}}_{{t}}^{{g}} ,
\end{equation}
\begin{equation}
\resizebox{\hsize}{!}{%
$
{Attn}_1(\mathcal{F}_{t}^{g}, \mathcal{F}_{v}) = \operatorname{Softmax}\left(\frac{(\mathcal{F}_{t}^{g} {W}_{Q^1}) (\mathcal{F}_{v} {W}_{K^1})^{T}}{\sqrt{d_{k}}}\right) (\mathcal{F}_{v} {W}_{V^1}),
$
}
\end{equation}
where $Attn_1$ uses a multi-head attention mechanism to perform the first layer of interaction, $\sigma _{{1}}$ denotes a learnable scaling coefficient for the residual connection, $W_{Q^1}$ projects the global text embeddings $\mathcal{F}_{t}^{g}$ into the query space, while $W_{K^1}$ and $W_{V^1}$ project the video features $\mathcal{F}_{v}$ into the key and value spaces, respectively. $\sqrt{d_k}$ is the scaling factor. 

Secondly, the video features $ \mathcal{F}_{{v}}$  serve as queries ${Q}_{{v}}$, while the global embeddings $\mathcal{F}_{{t}}^{{g}} $ function as keys $ {K}_{{t}}^{{g}}$ and values $ {V}_{{t}}^{{g}}$. It allows the visual features to focus on the most important descriptions in the text for the current action. This process yields the video features 
$\mathcal{F}_{{v}}^{g}$, which have been calibrated by the global context. It can be represented as:
\begin{equation}  
         \mathcal{F}_{{v}}^{g}=Attn_2(\mathcal{F}_{{v}}, \mathcal{F}_{{t}}^{{g*}})+\sigma _{{2}}\mathcal{F}_{{v}} ,
\end{equation}
\begin{equation}
\resizebox{\hsize}{!}{%
$
{Attn}_2(\mathcal{F}_{{v}}, \mathcal{F}_{{t}}^{{g*}}) = \operatorname{Softmax}\left(\frac{(\mathcal{F}_{{v}} {W}_{Q^2}) (\mathcal{F}_{{t}}^{{g*}} {W}_{K^2})^{T}}{\sqrt{d_{k}}}\right) (\mathcal{F}_{{t}}^{{g*}} {W}_{V^2}),
$
}
\end{equation}
where $Attn_2$ uses a multi-head attention mechanism to perform the second layer of interaction, $\sigma _{{2}}$ is the learnable coefficient balancing the update, $W_{Q^2}, W_{K^2}, W_{V^2}$ are the projection matrices specific to the second attention layer. Here, the video features $\mathcal{F}_{v}$ serve as the query source, while the updated global embeddings $\mathcal{F}_{t}^{g*}$ serve as the key and value source.

\subsubsection{Step-Aware Fusion Module}
\begin{figure}[t]
    \centering
         \includegraphics[width=0.9\linewidth]{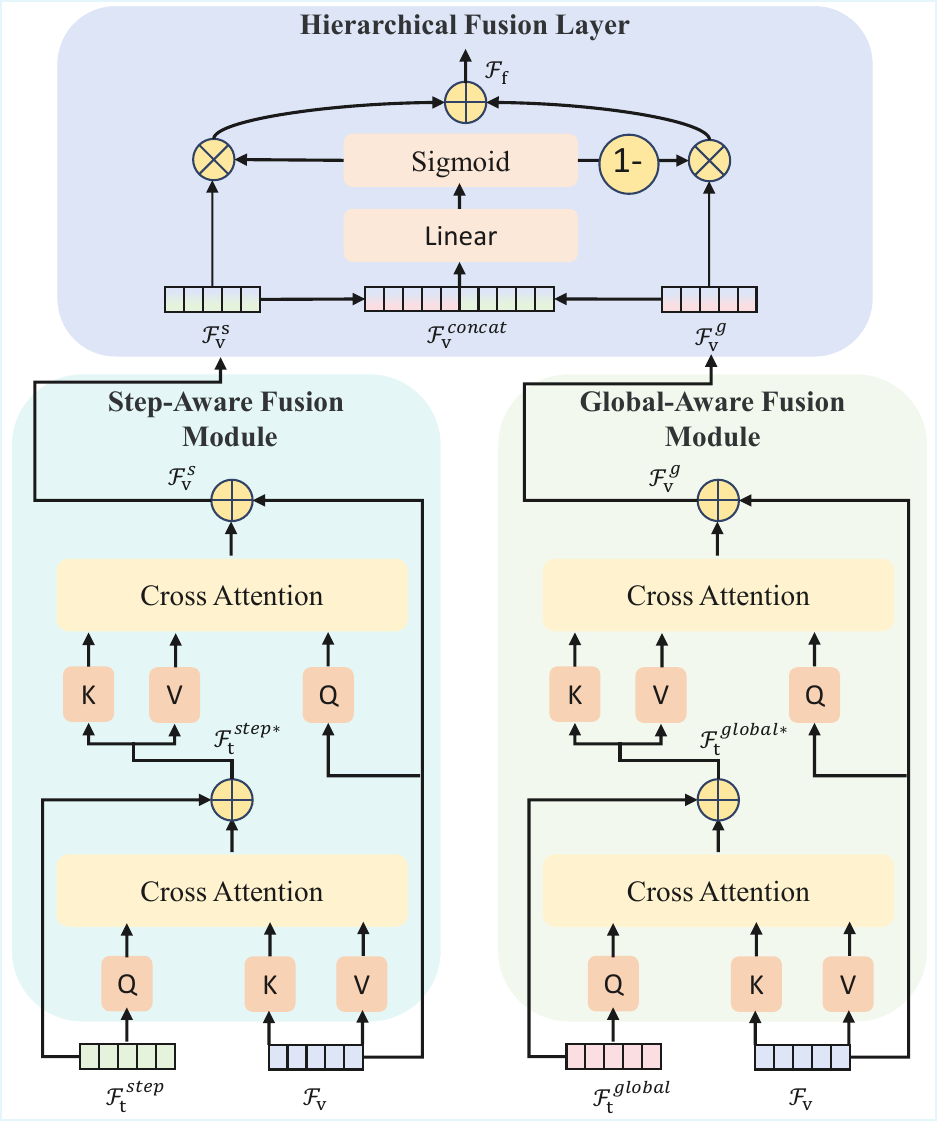}
         \vspace{-3mm}
    \caption{Illustration of the proposed Step-Aware Fusion Module, Global-Aware Fusion Module and Hierarchical Fusion Layer.}
		\label{figure_1}
		\vspace{-2mm}
	\end{figure}

Symmetrically, this module is designed to achieve fine-grained alignment between the visual features $ \mathcal{F}_{{v}}$ and the procedural step embeddings $ \mathcal{F}_{{t}}^{{s}}$. This enables the model to ground each specific step within the corresponding visual evidence. This module follows a two-part bidirectional interaction process, as shown in Figure \ref{figure_1}.

First, we employ the step embeddings $\mathcal{F}_{{t}}^{{s}}$  as queries $ {Q}_{{t}}^{{s}}$ and the video features $ \mathcal{F}_{{v}}$ as keys $ {K}_{{v}}$ and values $ {V}_{{v}}$ to ground each procedural step in the relevant video segments. This process can be represented as:
\begin{equation}  
         \mathcal{F}_{{t}}^{{s*}}= Attn_3 (\mathcal{F}_{{t}}^{{s}}, \mathcal{F}_{{v}})+\sigma _{{3}}\mathcal{F}_{{t}}^{{s}} ,
        \end{equation}
\begin{equation}
\resizebox{\hsize}{!}{%
$
{Attn}_3(\mathcal{F}_{{t}}^{{s}}, \mathcal{F}_{{v}}) = \operatorname{Softmax}\left(\frac{(\mathcal{F}_{{t}}^{{s}} {W}_{Q^3}) (\mathcal{F}_{{v}} {W}_{K^3})^{T}}{\sqrt{d_{k}}}\right) (\mathcal{F}_{{v}} {W}_{V^3}),
$
}
\end{equation}
where $Attn_3$ uses a multi-head attention mechanism to aligns fine-grained steps with visual content, $ \sigma _{{3}}$ represents a learnable coefficient. $W_{Q^3}$ maps the step embeddings $\mathcal{F}_{t}^{s}$ to the query, and $W_{K^3}, W_{V^3}$ map the video features $\mathcal{F}_{v}$ to the key and value.

Secondly, we employ the video features $ \mathcal{F}_{{v}}$ as the query ${Q}_{{v}}$, and the step embeddings $\mathcal{F}_{{t}}^{{s}} $  as the key $ {K}_{{t}}^{{s}}$ and value $ {V}_{{t}}^{{s}}$ to make the visual features aware of the specific procedural details. This process can be represented as:

 \begin{equation}  
         \mathcal{F}_{{v}}^{s}=Attn_4( \mathcal{F}_{{v}}, \mathcal{F}_{{t}}^{{s*}})+\sigma _{{4}}\mathcal{F}_{{v}} ,
        \end{equation}
\begin{equation}
\resizebox{\hsize}{!}{%
$
{Attn}_4(\mathcal{F}_{{v}}, \mathcal{F}_{{t}}^{{s*}}) = \operatorname{Softmax}\left(\frac{(\mathcal{F}_{{v}} {W}_{Q^4}) (\mathcal{F}_{{t}}^{{s*}} {W}_{K^4})^{T}}{\sqrt{d_{k}}}\right) (\mathcal{F}_{{t}}^{{s*}} {W}_{V^4}),
$
}
\end{equation}
where $Attn_4$ uses a multi-head attention mechanism to aligns fine-grained steps with visual content, $ \sigma _{{4}}$ represents a learnable coefficient, $W_{Q^4}$ projects the video features into queries to extract procedural details, while $W_{K^4}$ and $W_{V^4}$ project the updated step embeddings $\mathcal{F}_{t}^{s*}$ into keys and values.

Since then, we have obtained video features $\mathcal{F}_{{v}}^{g}$ that fuse the global text and video features $\mathcal{F}_{{v}}^{s}$ that fuse the step text.

\subsubsection{Hierarchical Fusion Layer}
This layer aims to combine the global-level fused features from one branch with the step-level fused features from the other. To achieve this, we employ a dynamic gating mechanism\cite{arevalo2017gated} that adaptively arbitrates the information flow from the two branches. As shown in Figure \ref{figure_1}, the fusion process begins by concatenating the two feature streams along their feature dimension. This combined representation provides the gating network with a complete view of both global and local perspectives at each spatio-temporal position. The gating network, composed of a linear layer and a Sigmoid activation function, is responsible for computing the gating tensor ${G}$. The entire dynamic gating and fusion process can be represented as:
\begin{equation}
{G} = \sigma({W}_{g}[\mathcal{F}_{{v}}^{g}, \mathcal{F}_{{v}}^{s}] + {b}_{g}),
\end{equation}
\begin{equation}
\mathcal{F}_{f} = {G} \odot \mathcal{F}_{{v}}^{g} + (1 - {G}) \odot \mathcal{F}_{{v}}^{s},
\end{equation}

\noindent where [*,*] denotes the concatenation operation, and the gating network is parameterized by a weight matrix ${W}_{g}$ and a bias vector  ${b}_{g}$. The sigmoid function $\sigma$ maps the output to a range of (0, 1), and the symbol $\odot$ represents the element-wise product.

\ar{ Traditional clip-based methods and standard cross-attention modules typically perform coarse-grained semantic matching or utilize straightforward feature addition/concatenation for fusion. However, in fitness coaching scenarios, evaluating a procedural step (\emph{e.g.}, ``bend the knees to 90 degrees") requires strict spatial-temporal causal reasoning. Standard fusion often suffers from spatial-temporal distraction, treating all frames and semantic tokens equally. Our dynamic gating mechanism systematically overcomes this limitation by acting as a kinematic filter: the explicit calculation of $G$ dynamically re-weights the step-level and global-level features at each specific spatio-temporal position. This structural design uniquely empowers the model to adaptively arbitrate between overarching action goals and fine-grained procedural steps, avoiding temporal noise and ensuring that specific visual frames are strictly aligned with their corresponding procedural instructions.}

Finally,  $ \mathcal{F}_{{f}}$ are fed into the prediction heads to yield the action class, quality assessment, and a Chain-of-Thought text explanation. Specifically, two independent MLPs classify the action class and quality, while the third component, a Transformer-based decoder \cite{lin2022swinbert}, auto-regressively produces the explanation token by token by conditioning on both the fused video-text features and previously generated words \cite{devlin2019bert}.

\subsection{Training and Inference}
During the training phase, we jointly optimize the entire framework in an end-to-end manner, enabling it to simultaneously learn three tasks: action classification, quality assessment, and CoT text generation. The total training loss is a weighted sum of three specialized losses: a multi-class Cross-Entropy loss $\mathcal{L}_c$ for action classification, a Binary Cross-Entropy loss $\mathcal{L}_q$ for quality assessment, and a language modeling loss $\mathcal{L}_t$ based on Cross-Entropy loss with label smoothing for CoT text generation:
 \begin{equation}  
         \mathcal{L} =\lambda\mathcal{L}_{{c}}+ \mathcal{L}_{{q}}+\mathcal{L}_{{t}} ,
        \end{equation}
where $\lambda$ is a scalar hyperparameter that balances the contribution of each task to the total loss. \ar{The initial search space for $\lambda$ was constrained by two primary factors: task dependency and loss magnitude. First, since accurate action classification establishes the upper bound for the subsequent quality assessment and explanation generation, its supervisory signal must be prioritized ($\lambda \ge 1$). Second, we empirically observed that the text generation loss ($\mathcal{L}_t$) inherently yields much larger numerical values than the classification loss ($\mathcal{L}_c$). Thus, scaling up $\mathcal{L}_c$ is mathematically necessary to prevent $\mathcal{L}_t$ from dominating the early gradient updates. Guided by these priors, we determined a broad initial screening range of $[0.5, 20]$, which is further thoroughly evaluated in Section \uppercase\expandafter{\romannumeral5}-C to justify our final selection of $\lambda=3$.}

During the inference stage, the model receives a video as input. It then simultaneously outputs the predicted action class and the quality assessment verdict. Based on the action class, it matches the corresponding standard technical steps to perform multimodal interaction, thereby auto-regressively generating a detailed Chain-of-Thought textual explanation.

In Algorithm \ref{alg}, we present the detailed training process for the EFA architecture. For each batch sample, we select the standard technical steps and the general instruction for the corresponding action category, input video frames (sampled at 6 frames per video) into a visual encoder to extract visual features, and simultaneously use a text encoder to extract text features. These unimodal features are fed into the Multimodal Fusion Module to generate fused features. Subsequently, two MLP classifiers predict action category and quality, while a Transformer-based text generator autoregressively generates chain-of-thought textual explanation from the fused features. The total model loss $ \mathcal{L} $ is a weighted sum of the action classification loss $ \mathcal{L}_c $, the action quality loss $ \mathcal{L}_q $ (balanced by weight $ \lambda $), and the text generation loss $ \mathcal{L}_t $. Finally, backpropagation is performed based on $ \mathcal{L} $, and the optimizer updates the model parameters $ \Theta $.

\begin{algorithm}[htb]
\color{black} %
\caption{The training process of EFA}
\label{alg}
\textbf{Input:} Training dataset $\mathcal{D} = \{ (V_i, y_i, q_i, t_i) \}_{i=1}^N$, where $V_i$ is the video, $y_i$ is the action class label, $q_i$ is the quality label, and $t_i$ is the Chain-of-Thought textual explanation. Learning rate $\eta$, loss weight $\lambda$, number of epochs $E$, batch size $B$.

\textbf{Output:} Model parameters $\Theta$
\begin{algorithmic}[1] 
\STATE Initialize: EFA model $M$ with parameters $\Theta$, Optimizer (\emph{e.g.}, AdamW).
\FOR{epoch = 1 to $E$}
    \FORALL{each batch $\{(V_i,  y_i, q_i, t_i)\}_{i=1}^B$ in $\mathcal{D}$}
        \STATE Retrieve standard technical steps and the general instruction $T_i = Map ( y_i)$
        \STATE  $\mathcal{F}_{v}  = \text{VisualEncoder}(V_i)$
        \STATE  $\mathcal{F}_{t}^{s},\mathcal{F}_{t}^{g} = \text{TextEncoder}(T_i)$
        \STATE  $\mathcal{F}_{f}  = \text{MultimodalFusionModule} (\mathcal{F}_{v}, \mathcal{F}_{t}^{s},\mathcal{F}_{t}^{g})$
        \STATE Predict action class $\hat{y}_i = \text{MLP}_c(\mathcal{F}_{f})$
        \STATE Predict quality label $\hat{q}_i = \text{MLP}_q(\mathcal{F}_{f})$
        \STATE \begin{tabular}[t]{@{}l@{}}
         Generate CoT text explanation $\hat{t}_i =$ \\
         $\text{TextGenerator}(\mathcal{F}_{f})$
       \end{tabular}
        \STATE  $\mathcal{L}_{total} = \mathcal{L}_c(\hat{y}_i, y_i) + \lambda \mathcal{L}_q(\hat{q}_i, q_i) + \mathcal{L}_t(\hat{t}_i, t_i)$ 
        \STATE Perform backpropagation on $\mathcal{L}_{total}$ and update parameters $\Theta$.
    \ENDFOR
\ENDFOR

\STATE \textbf{return}  $\Theta$.

\end{algorithmic}
\end{algorithm}

\section{EXPERIMENT}

\label{sec4}

\subsection{Experimental Setup}
\textbf{Evaluation Metrics.}
To comprehensively evaluate the performance of our proposed EFA and compare it with baselines, we employ a suite of standard evaluation metrics, covering both the text generation task for explainability and the classification tasks for quality and class prediction. For assessing the quality of the generated Chain-of-Thought text explanations, we utilize widely adopted metrics from text generation and video captioning: BLEU \cite{papineni2002bleu} measures the n-gram overlap to assess the precision of the generated sentences; METEOR \cite{banerjee2005meteor} considers synonym matching and stemming to better align the output with human perceptions of quality; CIDEr \cite{vedantam2015cider} uses TF-IDF weighting to emphasize the saliency and relevance of unique descriptive phrases; and ROUGE-L \cite{lin-och-2004-orange} focuses on the longest common subsequence to evaluate sentence-level fluency. \ar{Furthermore, to overcome the limitations of traditional N-gram metrics in capturing deep semantic reasoning, we introduce an advanced MLLM-as-a-Judge evaluation protocol using Qwen2.5-VL-72B. This protocol quantifies the generated feedback across three dedicated dimensions: Factuality, Logic, and Helpfulness. Specifically, Factuality measures the accuracy of identifying body movements without visual hallucinations. Logic assesses the soundness of the biomechanical causal reasoning chain. Finally, Helpfulness evaluates the practical utility and actionability of the corrective suggestions.} To evaluate the predictive capabilities of EFA, we use metrics tailored to each sub-task: for action class classification, we report the standard Top-1 and Top-5 accuracy metrics to assess the model's ability to correctly categorize the specific action from the 141 classes; and for the core binary task of action quality assessment, we report the overall classification Accuracy (Acc), which measures the percentage of correctly identified action forms.


\textbf{Implementation Details.}
We initialize the Text Encoder and Text Generator by utilizing a pre-trained BERT-base-uncased model \cite{devlin2019bert}. Separately, the Vision Encoder is initialized with a VideoSwin-Base network \cite{Liu_2022_CVPR} that has been pre-trained on Kinetics-600 \cite{Liu_2022_CVPR} to process 6 frames sampled from each video sequence. \ar{To rigorously justify this sampling choice and clarify the trade-off between task performance and computational overhead, a comprehensive ablation study across various frame rates (from 4 to 16 frames) is detailed in the supplementary material, empirically confirming that the 6-frame setting yields the optimal balance.} We split the dataset into training, test, and validation sets with proportions of 70\%, 15\%, and 15\%, respectively. We set both $\sigma_{{1}}$ and $\sigma_{{2}}$ to be a learnable parameter vector initialized to 1e-3, and set $\lambda$ = 3 . We employ the AdamW optimizer to train EFA  for 300 epochs, which follows a linear decay schedule with a warm-up phase. This process ran on 2 NVIDIA A6000 GPUs and consumed about 1 day. 

\subsection{Comparison with State-of-the-Art Methods}

\textbf{CoT Texts Generation.} As shown in Table \ref{exp1}, compared to the state-of-the-art approaches C3DAVG \cite{Parmar_2019_CVPR}, MSCADC \cite{Parmar_2019_CVPR},  SWINBERT \cite{lin2022swinbert}, LAVENDER \cite{li2023lavender} and RICA$^{2}$ \cite{majeedi2024ricaˆ} on CoT-AFA dataset, EFA achieved notable improvements on the four mainstream evaluation metrics for text generation BLEU \cite{papineni2002bleu}, METEOR \cite{banerjee2005meteor}, CIDEr \cite{vedantam2015cider} and ROUGE-L \cite{lin-och-2004-orange}. By outperforming previous methods by 4.5\%, 1.3\%, 16.0\%, and 4.7\% on the BLEU, METEOR, CIDEr, and ROUGE-L metrics respectively, our proposed model demonstrates its ability to recognize and understand the critical details and causal reasoning in video action sequences. The model has developed a capacity for textual generalization, eschewing superficial descriptions in favor of learning and utilizing expert-level vocabulary and phrases. The substantial 16.0\% improvement on the CIDEr score indicates that our EFA framework, guided by Standard Technical Steps, has learned to identify and articulate minor technical details with precise language (\emph{e.g.}, using ``lumbar hyperextension" instead of ``arching the back''). This indicates that EFA has moved beyond generic descriptions to provide expert-level analysis. In contrast, the more modest 1.3\% gain in METEOR can be attributed to the metric's inherent robustness, which inadvertently narrows the performance gap between our proposed model and baseline models in terms of terminological precision.  \ar{Beyond traditional metrics, our MLLM-as-a-Judge metrics (Table \ref{exp1}) explicitly validate EFA's deep semantic quality. By outperforming all baselines in Factuality, Logic, and Helpfulness, it robustly proves that EFA delivers not only grammatically correct texts, but also practically useful and biomechanically sound instructional feedback.}

\textbf{Action Classification and Quality Assessment.}
To validate the effectiveness of our proposed method in assessing action classification, we conduct experiments to compare it with action recognition methods. The results are reported in Table~\ref{exp2}, and we can find our proposed method achieves the best performance results compared to others and fully demonstrates the effectiveness of our approach in capturing global-local features and discriminate video differences.
For quality assessment, we re-classified the videos into only two categories for evaluation: standard and non-standard.
For action classification task, EFA achieves a Top-1 accuracy of 77.2\%, an improvement of 2.7\% over the strongest baseline (\emph{i.e.,} Video Swin \cite{Liu_2022_CVPR}). In the quality assessment task, our model obtains an 81.8\% accuracy, surpassing the best baseline by 2.1\%. These results validate our approach, demonstrating that incorporating textual guidance not only improves action recognition but also equips the model to capture the subtle visual details necessary for judging action form.

\begin{table*}[t] 
    \centering
    \caption{Comparison of the proposed EFA framework with baseline methods on AFA. The table reports performance using various metrics, including BLEU \cite{papineni2002bleu}, METEOR \cite{banerjee2005meteor}, CIDEr \cite{vedantam2015cider} and ROUGE-L \cite{lin-och-2004-orange} for text generation quality, \ar{and advanced MLLM-as-a-Judge metrics (Factual Correctness, Logical Consistency, Helpfulness) evaluated by Qwen2.5-VL-72B.} Higher values indicate better performance across all metrics.} 
    \label{exp1} 
    
        \begin{tabular}{@{}lcccccccc@{}}  
            \toprule
            \multirow{2}{*}{\textbf{Method}}  
            & \multicolumn{4}{c}{\textbf{N-gram Metrics}} 
            & \multicolumn{3}{c}{\textbf{MLLM-as-a-Judge Metrics }} \\ 
            \cmidrule(lr){2-5} \cmidrule(lr){6-8}
            & \textbf{BLEU} & \textbf{METEOR} & \textbf{CIDEr} & \textbf{ROUGE-L} & \textbf{Factuality} & \textbf{Logic} & \textbf{Helpfulness} \\ \midrule
            C3DAVG  & 39.4 & 18.8 & 16.1 & 30.3 & 45.24 & 48.50 & 42.31 \\
            MSCADC   & 37.9 & 17.6 & 15.8 & 29.8 & 44.78 & 48.08 & 42.48 \\
            SWINBERT   & 43.2 & 23.5 & 26.0 & 35.8 & 64.82 & 70.20 & 61.83 \\ 
            LAVENDER   & 47.1 & 22.8 & 28.7 & 36.0 & 62.31 & 67.44 & 61.87 \\
            RICA$^{2}$ & 44.4 & 21.0 & 27.3 & 33.8 & 53.25 & 57.11 & 50.88 \\ \midrule
            \textbf{EFA (Ours)} & \textbf{49.2}  & \textbf{23.8} & \textbf{33.3} & \textbf{37.7} & \textbf{68.00} & \textbf{72.81} & \textbf{64.69} \\ \bottomrule
        \end{tabular}%
    
\end{table*}

\begin{table}[t]
\centering
\caption{Comparison of our proposed method and existing action recognition methods (C3D \cite{Tran_2015_ICCV}, TSM \cite{Lin_2019_ICCV}, I3D \cite{Carreira_2017_CVPR}, Timesformer \cite{bertasius2021space} , Video Swin \cite{Liu_2022_CVPR}) and RICA$^{2}$ \cite{majeedi2024ricaˆ} on CoT-AFA. Action recognition (class) and action form assessment (quality). The best results are highlighted in bold.}  
  \small 
    \renewcommand{\arraystretch}{1.2}
    \setlength{\tabcolsep}{4.7mm}{
        \begin{tabular}{cccc} 
      \specialrule{0.1em}{1pt}{1pt} \hline
            \multirow{2}{*}{\textbf{Method}} 
            & \multicolumn{2}{c}{\textbf{Class}} 
            & \multicolumn{1}{c}{\textbf{Quality}} 
            \\ 
            & TOP1 & TOP5 
             & Acc 
            \\ 
            \hline
           
            C3D & 0.611  & 0.869  & 0.799   \\
            TSM & 0.723 & 0.932 & 0.707 \\
            I3D& 0.573 & 0.835 & 0.743  \\ 
            Timesformer & 0.735  & 0.951 & 0.793 \\
            Video Swin & 0.752  & 0.936 & 0.801 \\
            RICA$^{2}$ & 0.510 & 0.819 & 0.814 \\ \hline
           \textbf{ EFA(Ours) } & \textbf{0.772} & \textbf{0.959} & \textbf{0.818} \\
        \hline \specialrule{0.1em}{1pt}{1pt}
        \end{tabular}}
        \vspace{-2mm}

        \label{exp2}
\end{table}

\begin{table}[t]
    \centering
    \caption{Performance comparison of the full EFA model against ablated variants, demonstrating the effects of different components.} 
   
    {
    \renewcommand{\arraystretch}{1.3} 
    \setlength{\tabcolsep}{1pt} 
    \begin{tabular}{lcccccc}  
        \specialrule{0.1em}{1pt}{1pt} \hline
        \multirow{2}{*}{\textbf{Method}}  
        & \multicolumn{4}{c}{\textbf{Caption}} 
         & \textbf{Class}  
        & \textbf{Quality} 
        \\ 
        & BLEU & METEOR & CIDEr & ROUGE-L   & TOP1  & Acc
        \\            \hline
        \multicolumn{7}{l}{\textbf{1) Global/Step-Aware Fusion Module}}\\
        w/o Global-Aware Fusion & 48.8  &  23.3 & 32.4   & 35.0 & 0.760 & 0.817 \\
        w/o Step-Aware Fusion & 48.2  &  23.6 & 31.1 & 34.7 & 0.756 & 0.817 \\
        Q-text  & 48.0  &  23.1 & 29.0   & 36.9 & 0.753 & 0.818 \\
        Q-video  & 48.6 &  23.6 & 31.8   &37.0  & 0.770 & 0.802  \\ 
        \hline
        \multicolumn{7}{l}{\textbf{2) Hierarchical Fusion Layer}}\\
        Concatenate  & 48.8 &  23.6 & 31.4  & 37.4    & 0.764 & 0.803\\
        Add  & 48.6 &  23.7 & 33.3  & 36.1    & 0.759 & 0.805\\
        \hline
        \multicolumn{7}{l}{\textbf{3) Standard Technical Steps}}\\
        w/o text & 48.1  &  22.9 &  29.9  & 35.6 &  0.753 & 0.812 \\ 
        Shuffled text  & 49.0  & 23.1  &  30.3  &  37.6& 0.772& 0.811 \\
        \hline
       \textbf{EFA Full Model} &  \textbf{49.2}  & \textbf{23.8} & \textbf{33.3} & \textbf{37.7} & \textbf{0.772} &  \textbf{0.818}  \\
        \specialrule{0.1em}{1pt}{1pt} \hline
    \end{tabular}
    } 
    \vspace{-2mm}
    
    \label{exp3}
\end{table}

\begin{table}[ht]
    \centering
    \caption{Performance analysis with different values of the loss weight, demonstrating the trade-off in emphasizing the three tasks.} 
    
    {
    \renewcommand{\arraystretch}{1.2} 
    \setlength{\tabcolsep}{2.5pt} 
    \begin{tabular}{lcccccc}  
        \specialrule{0.1em}{1pt}{1pt} \hline
        \multirow{2}{*}{\textbf{Weight}}  
        & \multicolumn{4}{c}{\textbf{Caption}} 
         & \textbf{Class}  
        & \textbf{Quality} 
        \\ 
        & BLEU & METEOR & CIDEr & ROUGE-L   & TOP1  & Acc
        \\ \hline
        $\lambda = 0.5$ & 48.9  & 23.7 & 33.5 & 37.9 & 0.764 & 0.799 \\
        $\lambda = 1$   & 49.0  & 23.6 & 32.5 & 37.4 & 0.770 & 0.818 \\
        $\lambda = 3$   & \textbf{49.2} & \textbf{23.8} & \textbf{33.3} & \textbf{37.7} & \textbf{0.772} & \textbf{0.818} \\ 
        $\lambda = 5$   & 48.7  & 23.6 & 31.0 & 37.4 & 0.777 & 0.763 \\ 
        $\lambda = 10$  & 48.5  & 23.4 & 32.1 & 36.5 & 0.763 & 0.798 \\
        $\lambda = 15$  & 49.0  & 23.8 & 34.7 & 37.0 & 0.767 & 0.751 \\
        $\lambda = 20$  & 48.9  & 23.2 & 33.8 & 37.3 & 0.760 & 0.750 \\
        \specialrule{0.1em}{1pt}{1pt} \hline
    \end{tabular}
    } 
    \vspace{-2mm}
    
    \label{exp4}
\end{table}

\subsection{Ablation Study}
To systematically dissect the EFA framework and validate the contribution of each key component, we conducted a series of ablation studies. We primarily analyzed four core aspects: Dual-branch Global/Step-Aware Fusion Module, Hierarchical Fusion Layer, the quality of the Standard Technical Steps provided as input, and the value of the loss weight. All experiments were conducted on our collected CoT-AFA dataset.

\textbf{The effects of Global/Step-Aware Fusion Module.}
This study evaluates our proposed dual-branch fusion architecture and the necessity of its bidirectional attention mechanism. We tested several variants:
1) Removing Individual Branches: We individually removed the Global-Aware Fusion module (w/o Global-Aware Fusion) and the Step-Aware Fusion module (w/o Step-Aware Fusion). As shown in Table~\ref{exp3}, removing either branch leads to an identical and significant drop in performance. This confirms that both the high-level, global context and the fine-grained, step-by-step guidance are indispensable for the model's comprehensive understanding.
2) Removing Bidirectional Attention: We also compared our bidirectional interaction against two unidirectional variants: Q-text (only text queries video) and Q-video (only video queries text). The results show that while Q-video outperforms Q-text , our full EFA model with bidirectional attention still achieves the best performance. This strongly demonstrates that a mutual, bidirectional alignment strategy where visual features actively seek textual guidance ground themselves in visual evidence is critical for achieving the most effective multimodal fusion.

\textbf{The Effects of Hierarchical Fusion Layer.}
We examined the effectiveness of the Hierarchical Fusion Layer in integrating the outputs from the two parallel branches, and report the results in Table~\ref{exp3}. We replaced our proposed dynamic gating mechanism with two more basic fusion strategies: 1) \emph{Concatenate}: Direct feature concatenation. 2) \emph{Add}:  Element-wise addition. The results clearly indicate that these simpler approaches are less effective, as both methods result in lower scores. This highlights the advantage of the dynamic gating mechanism, which can dynamically regulate the information flow from the two branches, adaptively weighing the importance of global context versus specific step details to form a more robust final representation.

\textbf{The Impacts of Standard Technical Steps.}
To validate the role of structured textual guidance, we evaluated the impact of the quality of the Standard Technical Steps, and report the results in Table~\ref{exp3}. We compared our full model against two variants:
1) \emph{w/o text}: The model received no textual input and relied solely on visual information.
2) \emph{Shuffled text}: The standard technical steps were provided to the model in a randomly shuffled order.
As shown in the table, the quality of textual guidance is paramount for generating meaningful explanations. The complete absence of text (w/o text) severely degrades the model's descriptive capability, causing the CIDEr score to plummet to 29.9. Providing the text in a shuffled order (\emph{Shuffled text}) still results in a substantially inferior performance compared to our full EFA model. This result unequivocally demonstrates that the fine-grained, step-by-step technical instructions, provided in the correct sequence, are not just auxiliary information but are the key supervisory signal that empowers the model to perform causal reasoning and generate high-quality, instructive feedback.

\begin{figure*}[t]
    \centering
         \includegraphics[width=0.9\linewidth]{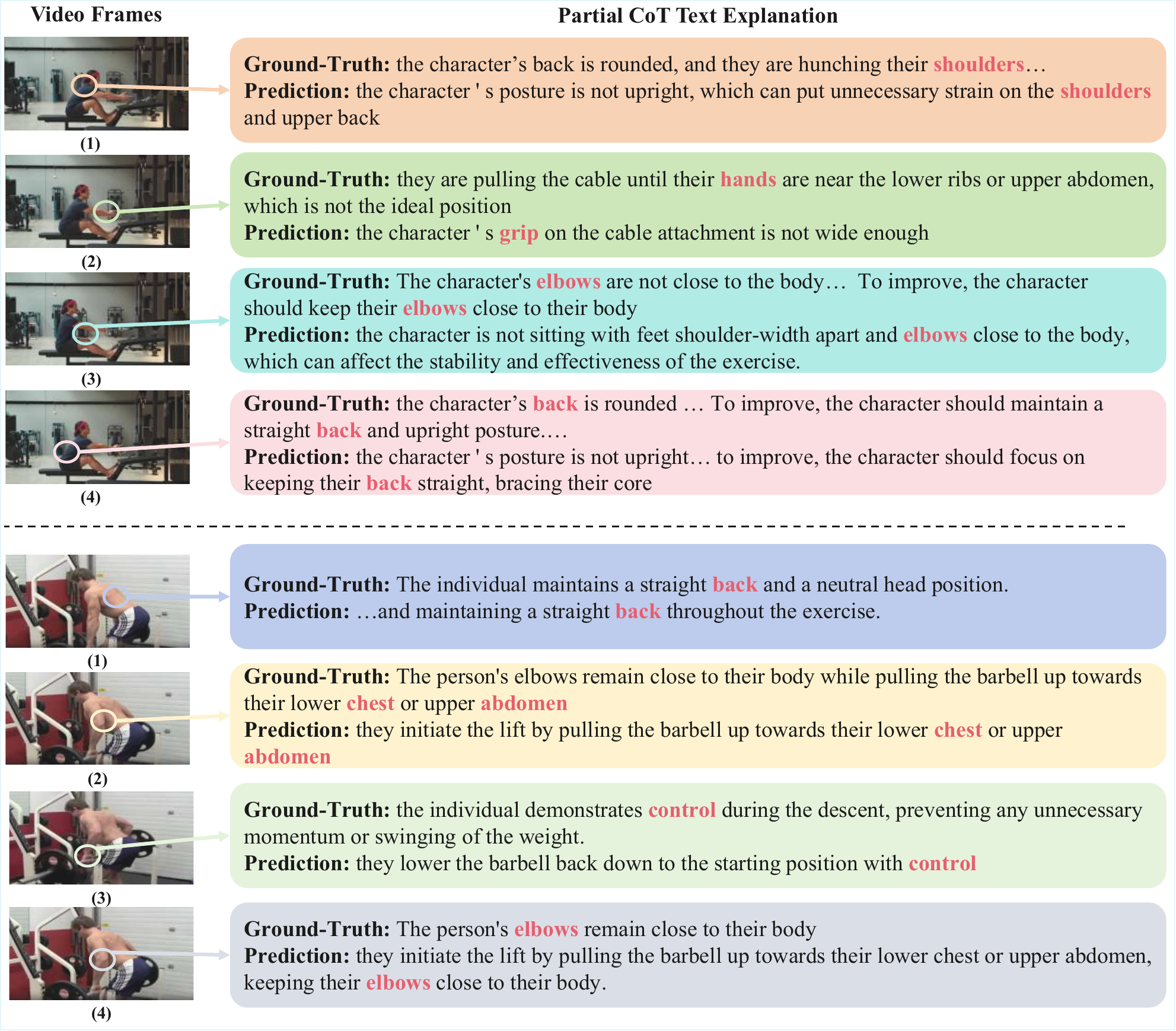}
         \vspace{-3mm}
    \caption{Examples of EFA's text generation for the non-standard "Seated Cable Row" action (top) and the standard "Barbell Bent-Over Row" action (bottom). Ground-Truth and Prediction are compared. Key techniques in the figure are marked and highlighted in the text with corresponding colors.}
		\label{figure4}
		\vspace{-1em}
	\end{figure*}

\textbf{The Effects of the Loss Weight.} To investigate the impact of the loss weight $\lambda$ on our multi-task loss function, we conducted a series of experiments by varying the weight assigned to the action classification loss $\mathcal{L}_{c}$ to observe the model's performance across all tasks. As shown in Table \ref{exp4}, when we set the weight to $\lambda = 3$, the model achieved the best or jointly best performance across all evaluation metrics. When the value of $\lambda$ deviated from 3, whether by decreasing it \ar{($\lambda \in \{0.5, 1\}$)} or increasing it \ar{($\lambda \in \{5, 10, 15, 20\}$)}, the model's overall performance showed varying degrees of decline. This suggests that providing \ar{an appropriately prioritized supervisory signal} for action classification helps the model learn more discriminative features, which in turn provides a better contextual foundation for generating accurate and relevant explanations. However, further increasing the weight to $\lambda$ = 5 and beyond leads to a slight degradation in performance, indicating that over-emphasizing the classification task may cause the model to neglect the fine-grained details necessary for quality assessment and text generation. Therefore, we conclude that $\lambda$ = 3 to provide the optimal balance.

\subsection{Visualization}

Figure \ref{figure4} provide several visual examples of the texts output from EFA (labeled Prediction), demonstrating its capability to perform both non-standard diagnosis and standard form confirmation based on the input video. For enhanced clarity and precise analysis, the CoT Textual Explanation has been segmentally broken down and aligned with the corresponding specific details observed in the individual video frames. The visualization presents two distinct scenarios. 

For the non-standard seated cable row example, EFA's prediction goes beyond simple fault detection, effectively employing the Chain-of-Thought paradigm. It details specific technical errors such as the character's grip on the cable attachment is not wide enough and the posture being not upright, which can put unnecessary strain on the shoulders and upper back. This causal explanation, linking the faulty technique to its potential negative consequence, is a key feature of  EFA framework. Furthermore, the model provides clear, actionable feedback, advising the user to focus on keeping their back straight, bracing their core to improve the form, directly fulfilling the explainability requirement of the AFA task.
Conversely, for the standard barbell bent-over row example, EFA generates CoT texts that accurately details the successful execution, validating key steps such as maintaining a straight back throughout the exercise, keeping their elbows close to their body, and performing the movement with control during the descent. This dual capability, identifying errors through logical reasoning while verifying correct form with detailed technical points, confirms that our multimodal fusion approach guided by the Standard Technical Steps, is highly effective. \ar{Furthermore, to provide a transparent understanding of the model's boundaries and limitations, we have conducted a dedicated failure and edge case analysis. Detailed discussions and qualitative visualizations of specific error modes, such as macro-level context absence and highly ambiguous borderline forms, are provided in the supplementary material.}

\ar{
\subsection{Robustness and Generalization Analysis}
To rigorously validate our framework's generalization, we conducted extensive out-of-distribution (OOD) and zero-shot experiments across three challenging protocols: cross-domain transfer to a novel clinical Rehabilitation Dataset, cross-dataset and cross-view robustness evaluations using an extended AQA-7 dataset, and cross-category generalization on 15 unseen actions. Our model consistently outperforms state-of-the-art baselines across all settings. Detailed configurations and comprehensive results for these analyses are provided in the supplementary material.
}

\ar{
\subsection{Inference Efficiency and Limitations}
Due to the autoregressive decoder, EFA's average inference latency is approximately 0.35 seconds per 6-frame sample. While this makes strict frame-by-frame real-time generation challenging, it fully supports ``action-level'' near real-time coaching. Since a typical fitness repetition takes 2-3 seconds, EFA seamlessly delivers immediate post-action feedback before the user initiates the next repetition. Future work will explore non-autoregressive decoding or model distillation to further minimize latency.
}


\section{CONCLUSION}
In this paper, we presented a new human action form assessment~(AFA) task and introduced a new, diverse yet challenging dataset~\emph{CoT-AFA} that has rich multi-element annotations and Chain-of-Thought text explanations. Furthermore, we proposed a new  Explainable Fitness Assessor framework to assess action form and provide explainability and detailed feedback. In experiments, the achieved state-of-the-art performance validates the effectiveness of our proposed model. In the future, we will apply the method in the robotic application.


\bibliographystyle{IEEEtran}
\bibliography{ref}

\end{document}